\documentclass[runningheads]{llncs}
\usepackage{epsfig}
\usepackage{graphicx}
\usepackage{amsmath}
\usepackage{amssymb}
\usepackage{float}
\usepackage{algorithm}
\usepackage[noend]{algpseudocode}

\graphicspath{{./Pictures/}}

\begin{document}
\title{Self-Net: Lifelong Learning via Continual Self-Modeling}
\titlerunning{Self-Net}
\author{Blake Camp\inst{1}\thanks{Both authors contributed equally.}\and
Jaya Krishna Mandivarapu\inst{1}$^{*}$\and
Rolando Estrada\inst{1}\orcidID{0000-0003-1607-2618}}
\authorrunning{Camp et al.}
\institute{Department of Computer Science, Georgia State University, Atlanta, GA 30303, USA\\
\email{\{bcamp2,jmandivarapu1\}@student.gsu.edu, restrada1@gsu.edu}\\
}
\maketitle              % typeset the header of the contribution
\begin{abstract}
% Continual learning (CL) is one of the most challenging problems in artificial intelligence. While several recent approaches achieve some degree of CL in deep neural networks, they are generally marred by unscalable storage requirements, inefficient training regimes,  or model saturation. In this paper, we present a scalable approach to continual learning that offers a practical solution to these problems. Motivated by the biological mechanisms responsible for consolidating knowledge and encoding experiences for long term storage, we present Self-Net, a novel framework which auto-encodes its own networks in a continual fashion.  We show that a modified contractive autoencoder can efficiently integrate entire networks into a compact latent space, and we demonstrate that the latent representations can be used to generate high-fidelity recollections of their original counterparts. The result is a single, compact model capable of generating an entire set of task-specific networks, each individually trained on a different task during the lifetime of the system. Our technique outperforms other state-of-the-art approaches on numerous datasets, including continual versions of MNIST, CIFAR10, CIFAR100, and Atari. To the best of our knowledge, we are the first to use autoencoders to sequentially encode entire sets of networks in order to facilitate continual learning.  

Learning a set of tasks over time, also known as continual learning (CL), is one of the most challenging problems in artificial intelligence. While recent approaches achieve some degree of CL in deep neural networks, they either (1) grow the network parameters linearly with the number of tasks, (2) require storing training data from previous tasks, or (3) restrict the network's ability to learn new tasks. To address these issues, we propose a novel framework, Self-Net, that uses an autoencoder to learn a set of low-dimensional representations of the weights learned for different tasks. We demonstrate that these low-dimensional vectors can then be used to generate high-fidelity recollections of the original weights. Self-Net can incorporate new tasks over time with little retraining and with minimal loss in performance for older tasks. Our system does not require storing prior training data and its parameters grow only logarithmically with the number of tasks. We show that our technique outperforms current state-of-the-art approaches on numerous datasets---including continual versions of MNIST, CIFAR10, CIFAR100, and Atari---and we demonstrate that our method can achieve over 10X storage compression in a continual fashion. To the best of our knowledge, we are the first to use autoencoders to sequentially encode sets of network weights to enable continual learning.

\keywords{Continual learning  \and Deep learning \and Autoencoders.}
\end{abstract}

\section{Introduction}
\label{sec:introduction}
Lifelong or continual learning (CL) is one of the most challenging problems in machine learning, and it remains a significant hurdle in the quest for artificial general intelligence (AGI) \cite{investigation_of_catastrophic_forgetting,measuring_catastrophic_forgetting}. In this paradigm, a single system must learn to solve new tasks without forgetting previously learned information. Different tasks might require different data (e.g., images vs. text) or they might process the same data in different ways (e.g., classifying an object in an image vs. segmenting it). Crucially, in CL there is no point at which a system stops learning; it must always be able to update its representation of its problem domain(s). 

% Here, a task refers to a desired mapping between inputs and outputs.

\begin{figure}[t]
	\centering
	\includegraphics[width=0.9\textwidth]{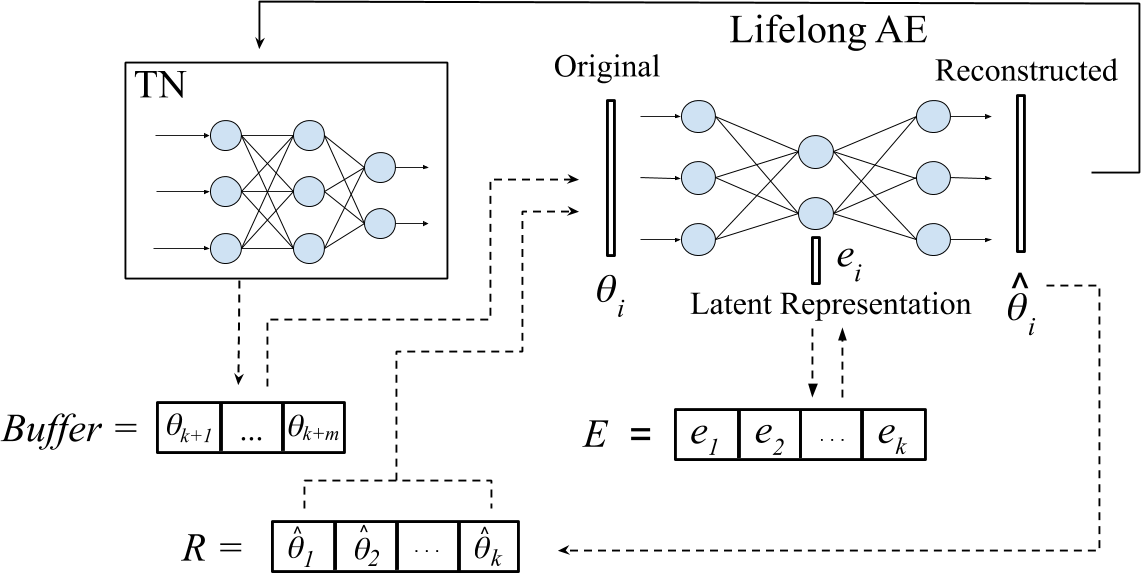}
	\vspace{-0.5em}
	\caption{\textbf{Framework overview:} Our proposed system has a set of reusable \textit{task-specific networks} (TN), a \textit{Buffer} for storing the latest $m$ tasks, and a lifelong, \textit{auto-encoder} (AE) for long-term storage. Given new tasks $\{t_{k+1}, ... ,t_{k+m}\}$, where $k$ is the number of tasks previously encountered, we first train $m$ task-networks to learn optimal parameters $\{\theta_{k+1},..., \theta_{k+m}\}$ for these tasks. These networks are temporarily stored in the Buffer. When the Buffer fills up, we incorporate the new networks into our long-term representation by retraining the AE on both its approximations of previously learned networks and the new batch of networks. When an old network is needed (e.g. when a task is revisited), we reconstruct its weights and load them onto the corresponding TN (solid arrow). Even when the latent representation $e_i$ is asymptotically smaller than $\theta_i$, the reconstructed network closely approximates the performance of the original.}
	\label{fig:overview}
	\vspace{-1em}
\end{figure}

CL is particularly challenging for deep neural networks because they are trained end-to-end. In standard deep learning we tune all of the network's parameters based on training data, usually via backpropagation \cite{Rumelhart1986}. While this paradigm has proven highly successful for individual tasks, it is not suitable for continual learning because it overwrites existing weights (a phenomenon evocatively dubbed \textit{catastrophic forgetting} \cite{robins1995catastrophic}). For example, if we first train a network on task A and then on task B, the latter training will modify the weights learned for A, thus likely reducing the network's performance on this task.

There are several approaches that can achieve some degree of continual learning in deep networks. However, existing methods suffer from at least one of three limitations: they either \textbf{(1)} restrict the network's ability to learn new tasks by penalizing changes to existing weights \cite{overcoming_catastrophic_forgetting,synaptic_intelligence,progress_and_compress,variational_continual_learning}; \textbf{(2)} expand the model size linearly as the number of tasks grows \cite{progressive_networks,LwF} (or dynamically define task-specific sub-networks \cite{dynamically_expandable_networks,context_dependent_gating}, which is asymptotically equivalent); or \textbf{(3)} retrain on old tasks. In the latter, we either \textbf{(a)} store some of the old training data directly \cite{playing_atari_deep_reinforcement,iCarl,variational_continual_learning}, thus increasing storage requirements linearly (and at a faster rate than increasing the network, since data tends to be higher dimensional), or \textbf{(b)} they use compressed data \cite{lifelong_generative_modeling,fearNet,deep_generative_replay,encoder_based_lifelong_learning}, which complicates training. 

%\cite{overcoming_catastrophic_forgetting, iCarl, progressive_networks, progress_and_compress, deep_generative_replay, synaptic_intelligence, uncompromising_incremental_learner, lifelong_generative_modeling, fearNet, variational_continual_learning, dynamically_expandable_networks, Deep_generative_dual_memory_for_CL, Gradient_episodic_memory, scalable_recollections, investigation_of_catastrophic_forgetting, measuring_catastrophic_forgetting, catastrophic_interference_causes_solutions_data,LwF, CL_through_evolvable_nerual_turing_machines,learn++, LL_dynamic_comination_model, lifelong_robot_learning, SupportNet, robust_evaluatons_of_continual_learrning, unicorn_CL_universal_off_policy_agent, CHILD, LL_of_humans_with_deep_neural_network_self_organization, context_dependent_gating, }

In this paper, we propose a novel approach, Self-Net, that overcomes the aforementioned limitations by decoupling how it \textit{learns} a new task from how it \textit{stores} it. Figure~\ref{fig:overview} provides an overview of our proposed framework. Our system grows only \textit{logarithmically} with the number of tasks, while retaining excellent performance across all learned tasks. Our approach is loosely inspired by the role that the hippocampus is purported to play in memory consolation \cite{hipp_memory_index_theory}. As noted in \cite{interplay_hippo_prefrontal_cortex}, during learning the brain forms an initial neural representation in cortical regions; the hippocampus then consolidates this representation into a form that is optimized for storage and retrieval. These complementary biological mechanisms enable continual learning by efficiently consolidating knowledge and prior experiences. In this spirit, we propose a system that consists of three components: \textbf{(1)} a set of reusable \textit{task-networks} (TNs), \textbf{(2)} a \textit{Buffer} in which we store the latest $m$ learned weights exactly, and \textbf{(3)} a lifelong \textit{autoencoder} (AE) with which we can encode an arbitrary number of older tasks. The AE learns a low-dimensional representation for each of the high-dimensional parameter vectors that define weights of the TNs. Thus, our system \textit{self-models} its own behavior, allowing it to approximate previously learned parameters instead of storing them directly. In short, when our system learns a new task, it firsts trains an appropriate TN using standard deep learning and then stores a copy of the weights in the Buffer. When the Buffer fill up, the AE learns a set of compact, latent vectors for the weights in the Buffer. The Self-Net then discards the original weights, freeing up the Buffer to store new tasks. If our system needs to solve a previously learned task, it generates an approximation of the original weights by feeding the corresponding latent vector through the AE and then loading the reconstructed weights onto a TN.

Our approach leverages the flexibility of conventional neural networks while avoiding their inability to remember old tasks. More specifically, a TN is free to modify its parameters as needed to learn a new task, since previously learned weights are encoded by the AE. Our AE doesn't simply memorize old weights; our experiments show that an AE can encode a very large number of networks while retaining excellent performance on all tasks (Section~\ref{sec:cl_large_tasks}). Our framework can even incorporate fine-tuning by initializing a TN with the weights from a previous, related task. Below, we first overview existing CL methods for deep network and then detail our approach in Section~\ref{sec:methodology}.

% The AE learns an $O(\log{(n)})$-representation of the $n$-dimensional parameter vector for each trained TN.

% , as described in detail in Section~\ref{sec:methodology}

% while retaining knowledge gained from previous tasks

% The result is a single model that can generate approximations of all previously-learned task-specific networks which retain comparable performance to their original counterparts.

% Overall, our proposed approach learns multiple consecutive tasks in a sequential manner without forgetting old tasks.

% in contrast to several competing techniques, we place only mild restrictions on the task-networks as they update parameters during training.  This is permissible and encouraged because the parameters themselves are never explicitly saved, but are rather encoded and approximated by the AE.  As a result, it matters not if the original learned parameters for task $t_1$ are overwritten during training of task $t_2$.  Further, this allows the framework to leverage the benefits of fine-tuning by initializing the task-network with the weights from a previous task. 

% Additionally, our approach is not limited to using a single architecture; as we validate in our experiments (Section~\ref{sec:experiments}), our AE can simultaneously store architectures of multiple types and sizes within the same latent space. It also allows us to efficiently store large numbers of networks. 

\section{Prior work}
\label{sec:priorWork}
Several methods have recently emerged for continual learning in deep networks, although, as noted above, existing approaches either \textbf{(1)} restrict new learning, \textbf{(2)} grow the number of parameters linearly, or \textbf{(3)} require old training data. Notable examples of the first type include Elastic Weight Consolidation (EWC) \cite{overcoming_catastrophic_forgetting}, Synaptic Intelligence \cite{synaptic_intelligence}, Variational Continual Learning \cite{variational_continual_learning} (which also reuses old data), and Progress \& Compress \cite{progress_and_compress}. These approaches reuse the same network for each new task, but they apply a regularization method to restrict changes in weights over time. Hence, they typically use constant space\footnote{Although, as noted in \cite{HuszarE2496}, standard EWC stores an $O(n)$ set of Fisher weights for each task, so it actually grows linearly. The modified version proposed in \cite{HuszarE2496} does use constant space.}. EWC, in particular, uses the diagonal of the Fisher information matrix between the weights learned for the new task vs. the old tasks. Like our proposed approach, Progress \& Compress also uses both a task-network and a long-term storage network; however, it uses EWC to update the weights of the latter, so it has very similar performance to this first method.

The second category includes Progressive Networks \cite{progressive_networks}, Dynamically Expandable Networks \cite{dynamically_expandable_networks}, and Context-Dependent Gating \cite{context_dependent_gating}. These methods achieve excellent performance, but they grow the network linearly with the number of tasks, which is asymptotically the same as using independent networks. Thus, they cannot scale to large numbers of tasks. Their advantage is in facilitating \textit{transfer learning}, i.e., using previous learning to speed up new learning. 

Finally, some methods store a fraction of the old training data and use it to retrain the network on previously learned tasks. Key approaches include Experience Replay \cite{playing_atari_deep_reinforcement} iCarl \cite{iCarl}, Variational Continual Learning \cite{variational_continual_learning}, and Learning without Forgetting \cite{LwF}. Unfortunately, this paradigm combines the drawbacks of the previous two. First, most of these methods use a single network, so they cannot continually learn a large number of tasks well. Second, their storage requirements grow linearly in the number of tasks because they have to store old training data. Moreover, data usually takes up orders of magnitude more space than the network itself because a trained network is effectively a compressed representation of the training set \cite{2016arXiv160605908D}. A few methods reduce this storage requirement by storing a compressed representation of the data. Methods of this type include Lifelong Generative Modeling \cite{lifelong_generative_modeling}, FearNet \cite{fearNet}, and Deep Generative Replay \cite{deep_generative_replay}. Our proposed approach uses a similar idea but instead stores the \textit{networks themselves}, rather than the data. Our scheme has two advantages over compressing the data. First, networks are much smaller, so we can encode them more quickly, using less space. Second, by reconstructing the networks directly, we do not need to retrain task-networks on data from previous tasks.

% 	Given a new task $t_{k+1}$, we first train a TN to learn a set of parameters $\theta_{k+1}$ for this task. We then incorporate the new parameters $\theta_{k+1}$ into our long-term representation by retraining the AE on both its approximations of previously learned networks as well as the new network. 

\section{Methodology}
\label{sec:methodology}
Figure~\ref{fig:overview} provides a high-level overview of our proposed approach. Our Self-Net system uses a set of reusable task-networks (TNs), a Buffer for storing newly learned tasks, and a lifelong autoencoder (AE) for storing older tasks. In addition, we store an $O(\log{(n)})$ latent vector for each task. Each TN is just a standard neural network, which can learn regression, classification, or reinforcement learning tasks (or some combination of the three). For ease of discussion, we will focus on the case where there is a single TN and the Buffer can hold only one network; the extension to multiple networks and larger Buffers is trivial. The AE is made up of an \textit{encoder} that compresses an input vector into a lower-dimensional, latent vector $e$ and a \textit{decoder} that maps $e$ back to the higher-dimensional space. Our system can produce high-fidelity recollections of the learned weights, despite this intermediate compression. In our experiments, we used a contractive autoencoder (CAE) \cite{CAE} due to its ability to quickly incorporate new values into its latent space.

% The AE learns a low-dimensional representation of the high-dimensional parameter vectors that encode each learned task. 

In CL, we must learn $k$ different tasks sequentially. To learn these tasks independently, one would need to train and save $k$ networks, with $O(n)$ parameters each, for a total of $O(kn)$ space. In contrast, we propose using our AE to encode each of these $k$ networks as an $O(\log(n))$ latent vector. Thus, our method uses only $O(n + k\log{(n)})$ space, where the $O(n)$ term accounts for the TNs and the fixed-size Buffer. Despite this compression, our experiments show that we can obtain a high-quality approximation of previously learned weights, even when the number of tasks exceeds the number of parameters in the AE (Sec.~\ref{sec:cl_large_tasks}). Below, we first describe how to encode a single task-network before discussing how to encode multiple tasks in a continual fashion.

% Our method also use an $O(n)$ Buffer for recent tasks, but its size is a constant. Thus, in the asymptotic case where $k >> m$, our approach uses only $O(k\log(n))$ space.

\subsection{Single-network encoding}
Let $t$ be a task (e.g., recognizing faces) and let $\theta$ be the $O(n)$-dimensional vector of parameters of a network trained to solve $t$. That is, using a task-network with parameters $\theta$, we can achieve performance $p$ on $t$ (e.g., a classification accuracy of 95\%). Now, let $\hat{\theta}$ be the approximate reconstruction of $\theta$ by our autoencoder and let $\hat{p}$ be the performance that we obtain by using these reconstructed weights for task $t$. Our goal is to minimize any performance loss w.r.t. the original weights. If the performance of the reconstructed weights is acceptable, then we can simply store the $O(\log{(n)})$ latent vector $e$, instead of the $O(n)$ original vector $\theta$.

If we had access to the test data for $t$, we could assess this difference in performance directly and train our AE until we achieve an acceptable margin $\epsilon$:
\begin{align}
p - \hat{p} \leq \epsilon.
\end{align}
For example, for a classification task we could stop training our AE if the drop in accuracy is less than $1\%$. 

% , which is infeasible for large numbers of tasks

In a continual learning setting, though, the above scheme requires storing validation data for each old task. Instead, we measure a distance between the original and reconstructed weights and stop training when we achieve a suitably close approximation. Empirically, we determined that the cosine similarity,
\begin{align}
\label{cos_sim}
cos(\theta, \hat{\theta}) = \dfrac{\theta \cdot \hat{\theta}}{\|\theta\|  \|\hat{\theta}\|} = \dfrac{\sum_{i=1}^{n}\theta_i \hat{\theta}_i}{\sqrt{\sum_{i=1}^{n}\vphantom{\hat{\theta}_i^2} \theta_i^2 } \sqrt{\sum_{i=1}^{n}\hat{\theta}_i^2}},
\end{align}
is an excellent proxy for a network's performance. Unlike the mean-squared error, this distance metric is scale-invariant, so it is equally suitable for weights of different scales. As detailed in Section~\ref{sec:experiments}, a cosine similarity of 0.997 or higher yielded excellent performance for a wide variety of tasks and architectures. 

In addition, one can improve the efficacy with which the AE learns a new task by encouraging the parameters of all task-networks to remain in the same general neighborhood. This can be accomplished by fine-tuning all networks from a common source and penalizing large deviations from this initial configuration with a regularization term.  Formally, let $\theta^{*}$ be the source parameters, ideally optimized for some highly-related task.  Without loss of generality, we can define the loss function of task-network $\theta_{i}$ for task $t_{i}$ as:
\begin{align}
\label{eqn:regularized_task_loss}
TaskNetLoss_{i} = TaskLoss +  \lambda  MSE(\theta^{*}, \theta_{i})
\end{align}
where $\lambda$ is the regularization coefficient determining the importance of remaining close to the source parameters vs. optimizing for the current task.  By encouraging the parameters for all task-networks to remain close to one another, we make it easier for the AE to learn a low-dimensional representation of the original space.

% manifold which can effectively model the original high-dimensional parameter space.

\subsection{Continual encoding}
\label{sec:continual-encoding}
We will now detail now to use our Self-Net to encode a sequence of trained networks in a continual fashion. Let $m$ be the size of the Buffer, and let $k$ be the number of tasks which have been previously encountered. As noted above, we train each of these $m$ task-networks using conventional backpropagation, one per task. Now, assume that our AE has already learned to encode the first $k$ task-networks. We will now show how to encode the most recent batch of $m$ task-networks corresponding to tasks $\{t_{k+1}, ... ,t_{k+m}\}$ into compressed representations $\{e_{k+1}, ... ,e_{k+m}\}$ while still remembering all previously trained networks. 

\begin{algorithm}[t]
{\footnotesize
	\caption{Lifelong Learning via Continual Self-Modeling}\label{alg:mccl}
	\begin{algorithmic}[1]
		
		\State Let \textbf{\textit{T}} be the set of all Tasks encountered during the lifetime of the system
		\State Let $m$ be the size of the \textbf{Buffer}
		\State \textbf{\textit{E}} = []  
		\State initialize \textbf{AE}
		\State Set cosine\_threshold
		
		\For {idx,curr\_task in enumerate(\textbf{\textit{T}})}
    		\If {\textbf{Buffer} is \textbf{not} full}

        		\State - Intitialize \textbf{TN} 
        		
         		\State - Train the \textbf{TN} for curr\_task until optimized 
         		\State - \textbf{Buffer}.append(\textbf{TN})

         		\If {\textbf{Buffer} \textbf{is} full}

             		\State \textbf{\textit{R}} = []
            		
            		\For{ encoded-network in \textbf{\textit{E}}}
                		\State r = \textbf{AE}.Decoder(encoded-network)
                		\State \textbf{\textit{R}}.append(r)
            		\EndFor
            		
            		\For{ network in \textbf{Buffer}}
                		\State flat\_network = extract and flatten parameters from network
                		\State \textbf{\textit{R}}.append(flat\_network)
            		\EndFor
            		
            		\State average\_cosine\_similarity = $0.0$
            		\State \textbf{\textit{E}} = []
            		\While { average\_cosine\_similarity $<$ cosine\_threshold}
                		\For {r\_idx,\textbf{r} $\in$ enumerate(\textbf{\textit{R}})}
                    		\State calculate AE\_loss using Equation \eqref{eqn:caeLoss}.
                    		\State back-propagate \textbf{AE} w.r.t \textbf{r}
                    		\State update average\_cosine\_similarity using cos(\textbf{r},\textbf{AE}(\textbf{r}))
                    		\State \textbf{\textit{E}}[r\_idx] = \textbf{AE}.Encoder(\textbf{r})
                		\EndFor
            		\EndWhile
            		\State empty \textbf{Buffer}
         		\EndIf
     		\EndIf
		\EndFor
		
	\end{algorithmic}
}
	\label{alg:csm_algo}
\end{algorithm}

% As noted above, we first learn parameters $\{\theta_{k+1}, ..., \theta_{k+m}\}$ by training $m$ task-networks until we achieve suitable performances across tasks $\{t_{k+1},...,t_{k+m}\}$. In our experiments, this step consists of conventional ANN training, i.e., backpropagation w.r.t. to large training sets, one per task. We then extract and flatten the learned parameters into $n$-dimensional vectors, where $n$ is the total number of parameters in the network. Networks 

Let $E$ be the set of latent vectors for the first $k$ networks. In order to integrate $m$ new networks $\{\theta_{k+1}, ..., \theta_{k+m}\}$ into the latent space, we first recollect all previously trained networks by feeding each $e \in E$ as input to the decoder of the AE. We thus generate a set $R$ of recollections, or approximations, of the original networks (see Fig.~\ref{fig:overview}). We then append each network $\theta_{i}$ in the Buffer to $R$ and retrain the AE on all $k+m$ networks until it can reconstruct them, i.e., until the average of their respective cosine similarities is above the predefined threshold. Algorithm \ref{alg:csm_algo} summarizes our continual learning strategy.

As we show in our experiments, our compressed network representations still achieve excellent performance compared to the original parameters. Since each $\hat{\theta} \in R$ is simply a vector of network parameters, it can easily be loaded back onto a task-network with the correct architecture. This allows us to discard the original networks and store $k$ networks using only $O(k\log{(n)})$ space. In addition, our framework can efficiently encode many different types and sizes of networks in a continual fashion. In particular, we can encode a network of arbitrary size $q$ using a constant-size AE (that takes inputs of size $n$) by splitting the input network into $r$ subvectors\footnote{We pad with zeros whenever $q$ and $n$ are not multiples of each other.}, such that ($n = q/r$). As we verify in Section~\ref{sec:experiments}, we can effectively reconstruct a large network from its subvectors and still achieve a suitable performance threshold.

\begin{figure}[t]
    \centering
	\includegraphics[ width=0.9\linewidth]{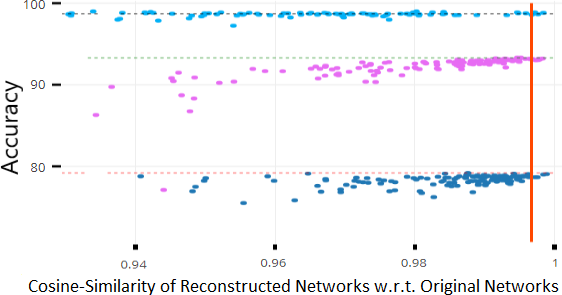}
% 	\vspace{-0.5em}
	\caption{\textbf{Robustness analysis of network performance as a function of cosine similarity:} Each dot represents the accuracy of a reconstructed network and the dotted lines are the baseline performances of the original networks. The above values for three datasets (Permuted MNIST (in pink), MNIST (in cyan), and CIFAR-10 (in blue), show that cosine similarity values above 0.997 guarantee nearly optimal performance.}
	\label{fig:robustness_combined}
% 	\vspace{-1em}
\end{figure}

As Fig.~\ref{fig:robustness_combined} illustrates, we empirically found a strong correlation between a reconstructed network's performance and its cosine similarity w.r.t. to the original network. Intuitively, this implies that vectors of network parameters that have a cosine similarity approaching 1 will exhibit near-identical performance on the underlying task. Thus, the cosine similarity can be used as a terminating condition during retraining of the AE. That is, there exists a cosine similarity threshold above which the performance of the reconstructed network can be expected to be sufficiently similar to that of the original. In practice, we found a threshold of .997 to be sufficient for most experiments. Below, we offer empirical results which demonstrate the efficacy and flexibility of our approach.

\section{Experimental Results}
\label{sec:experiments}
In order to evaluate the continual-learning performance of Self-Net, we carried out a range of experiments on a variety of datasets, in both supervised and reinforcement-learning (RL) settings. We first performed a robustness analysis to establish the degree to which an approximation of a network can deviate from the original and still retain comparable performance on the underlying task (Section~\ref{sec:robust}). Then, we evaluated the performance of our approach on the following continual-learning datasets:  Permuted MNIST \cite{overcoming_catastrophic_forgetting}, Split MNIST \cite{variational_continual_learning}, Split CIFAR-10 \cite{synaptic_intelligence}, Split CIFAR-100 \cite{synaptic_intelligence}, and successive Atari games \cite{playing_atari_deep_reinforcement} (we describe each dataset below). As our experiments show, Self-Net can effectively encode each of these different types of networks in sequential fashion, effectively achieving continual learning and outperforming several competing techniques. Finally, we also analyzed our system's performance under three additional scenarios: \textbf{(1)} very large numbers of tasks, \textbf{(2)} different sizes of AEs, and \textbf{(3)} different task-network architectures. We detail each experiment below.

\subsection{Robustness analysis}
\label{sec:robust}
Our approach relies upon \textit{approximations} of previously learned networks, and we assume no access to validation data for previously learned tasks. Thus, we require a method for estimating the performance of a reconstructed network which does not rely upon explicit testing on a validation set. 

Figure \ref{fig:robustness_combined} shows the relationship between performance and deviations from the original parameters as measured by cosine similarity, for three datasets. There is a clear correlation between the amount of parameter dissimilarity and the probability of a decrease in performance.  That is, given an approximate network that deviates from the original by some amount, the potential still exists that such a network will retain comparable performance.  However, as the degree of deviation increases, the probability that the performance remains high falls steadily. Thus, in order to assume, with reasonable confidence, that the performance of a reconstructed network will be sufficiently high, the AE must minimize the degree of deviation as much as possible. 

Empirically, we established a cosine similarity threshold above which the probability of high task-performance stabilizes, as seen in Figure \ref{fig:robustness_combined}.  This threshold can be used as a terminating condition during retraining of the AE, and it allows the performance of a reconstructed network to be approximated \textit{without} access to any validation data.  In our experiments, a common threshold yields good performance across a variety of different types and sizes of networks.

% , though it is sensitive to the size of the AE, as explained in Section \ref{sec:cl_large_tasks}.

% \subsection{Batch learning}
% \label{sec:batch_learning}
% We also verified that our AE could encode a large set of learned networks using conventional, batch learning, as opposed to sequential or continual learning. Figure \ref{fig:mnist_333_batch} shows the mean and SD of the performances of 333 approximated networks as a function of how long we train the AE.  The original networks were trained on 333 distinct MNIST classification tasks, each achieving approximately 98\% accuracy on a validation set.  The results demonstrate that, even for hundreds of tasks, the mean accuracy of the reconstructed networks is quite high, almost perfectly matching the mean accuracy of the original networks.

% \begin{figure}[H]
% \begin{center}

%   \includegraphics[width=0.99\linewidth]{mnist_333_batch.png}
% \end{center}
%   \caption{Mean Accuracy and Standard-deviation for 333 reconstructed MNIST networks trained as a batch. (i.e. non-continual learning)}
% \label{fig:mnist_333_batch}
% \end{figure}

\subsection{Experiments on CL datasets}
\label{sec:cl_experiments}
\begin{figure}
    \centering
	\includegraphics[width=.9\linewidth]{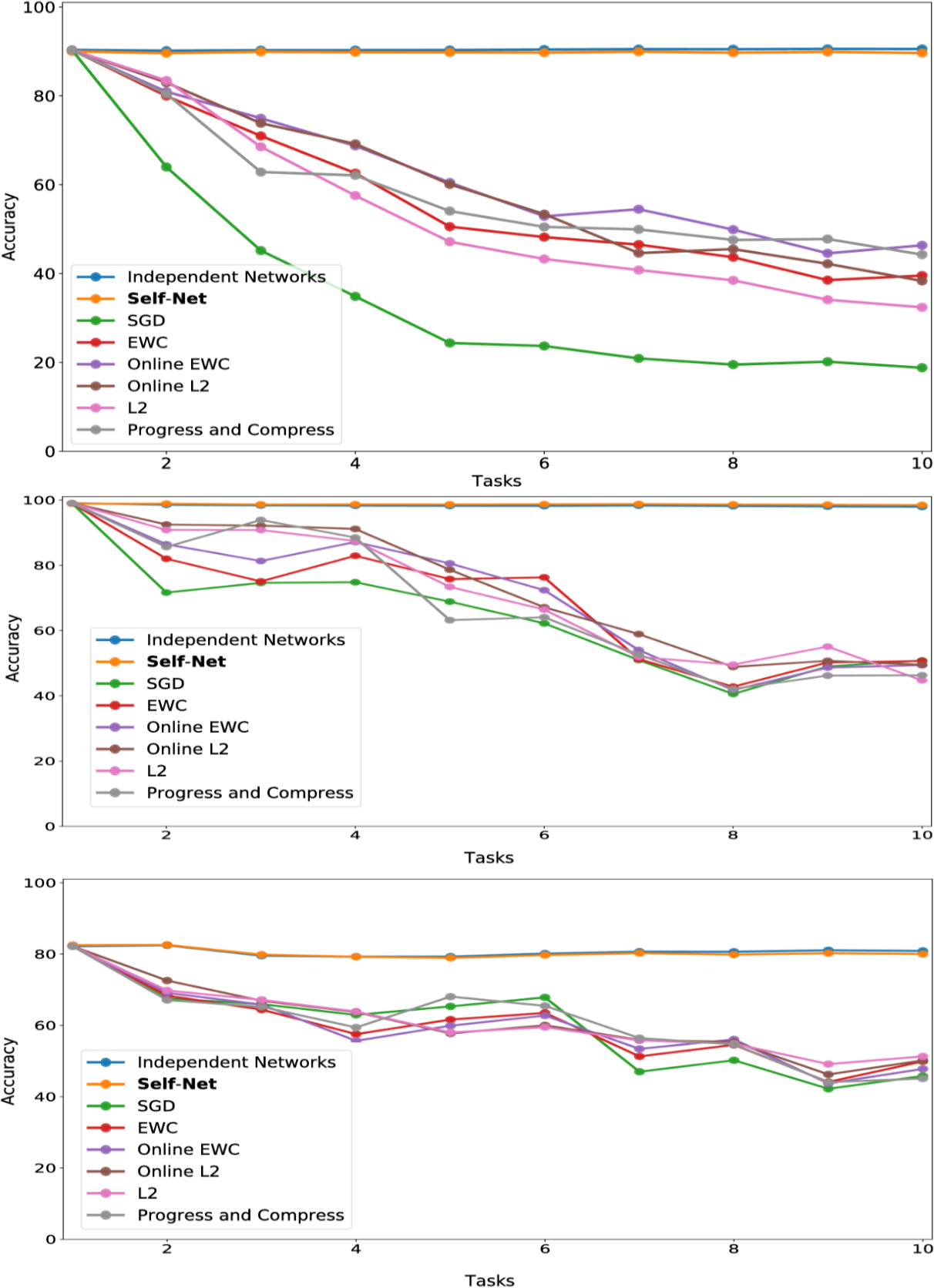}
% 	\vspace{-0.5em}
	\caption{CL performance comparisons with average test set accuracy on all observed tasks at each stage for \textbf{(top)} Permuted MNIST, \textbf{(middle)} Split MNIST, and \textbf{(bottom)} Split CIFAR-10.}
	\label{fig:combined_comparisons}
% 	\vspace{-1em}
\end{figure}

% \label{sec:cl_pmnist}
\textbf{Permuted MNIST}: As an initial evaluation of Self-Net's CL performance, we trained convolutional feed-forward neural networks with 21,840 parameters on successive tasks, each defined by distinct permutations of the MNIST dataset \cite{726791}, for 10-digit classification. We used networks with 2 convolution layers (kernels of size 5x5, and stride 1x1), 1 hidden layer (320x50), and 1 output layer (50x10). Our CAE had three, fully connected layers with 21,840, 2000, and 20 parameters, resp. Thus, our latent vectors were of size 20. For this experiment, we used a Buffer of size 1.  Each task network was encoded by our lifelong AE in sequential fashion, and the accuracies of all reconstructed networks were examined at the end of each learning stage (i.e., after learning a new task). Figure \ref{fig:combined_comparisons} (top) shows the mean performance after each stage. Our technique almost perfectly matched the performances achieved by independently trained networks, and it dramatically outperformed other state-of-the-art approaches including EWC \cite{overcoming_catastrophic_forgetting}, Online EWC (the correction to EWC proposed in \cite{HuszarE2496}), and Progress \& Compress \cite{progress_and_compress}. As a baseline, we also show the results for SGD (no regularization), L2-based regularization in which we compare new weights to all the previous weights, and Online L2, which only measures deviations from the weights learned in the previous iteration. Not only does our technique allow for superior knowledge retention, but it does not inhibit knowledge acquisition necessary for new tasks. The result is minimal degradation in performance as the number of tasks grow.

% \begin{figure}[H]
% \begin{center}

%   \includegraphics[width=0.99\linewidth]{mean_accuracy_Permuted_MNIST_final.pdf}
% \end{center}
%   \caption{Continual Learning on Permuted MNIST tasks.  Lines shown denote the mean accuracies, for each technique, across all tasks at successive stages of learning.}
% \label{fig:permuted_mnist}
% \end{figure}

% Further, as seen in Figure \ref{fig:pmnist_updates}, the total number of updates required to integrate a new network into the latent space of the lifelong AE is quite small (often a few hundred) when compared to the number of updates needed to learn the original task using task-data (usually many thousands, or even millions).  

% \begin{figure}[H]
% \begin{center}

%   \includegraphics[width=0.99\linewidth]{pmnist_updates.png}
% \end{center}
%   \caption{Total number of updates required by AE to accurately reconstruct all networks trained on Permuted MNIST tasks.  Specifically, the plots show the number of updates required to hit the cosine-similarity threshold for all networks.}
% \label{fig:pmnist_updates}
% \end{figure}

% \subsection{Split MNIST}
% \label{sec:incremental_mnist}
\textbf{Split MNIST:} We performed a similar continual learning task but with different binary classification objectives on subsets of the MNIST dataset (Split MNIST) \cite{variational_continual_learning}.  Our task-networks, CAE, and Buffer size were the same as for Permuted MNIST (except that the outputs of the task-networks were binary, instead of 10 classes). Tasks were defined by tuples comprised of the positive and negative digit class(es), e.g., ([pos=\{1\}, neg=\{6,7,8,9\}], [pos=\{6\}, neg=\{1,2,3,4\}], etc.). Here, the training and test sets consisted of approximately 40\% positive examples and 60\% negative examples. In this domain, too, our technique dramatically outperformed competing approaches, as seen in Figure \ref{fig:combined_comparisons} (middle).

% \begin{figure}[H]
% \begin{center}

%   \includegraphics[width=0.99\linewidth]{mean_accuracy_Incremental_MNIST_final.pdf}
% \end{center}
%   \caption{Continual learning performances on Incremental MNIST tasks}
% \label{fig:incremental_mnist}
% \end{figure}

% \subsection{Split CIFAR-10}
% \label{sec:cifar_10}
\textbf{Split CIFAR-10:} We then verified that our proposed approach could reconstruct larger, more sophisticated networks. Similar to the Split MNIST experiments above, we divided the CIFAR-10 dataset \cite{krizhevsky2009learning} into multiple training and test sets, yielding 10 binary classification tasks (one per class). We then trained a task-specific network on each class. Here, we used TNs having an architecture which consisted of 2 convolutional layers, followed by 3 fully connected hidden layers, and a final layer having 2 output units. In all, these task networks consisted of more than 60K parameters. Again, for this experiment we used a Buffer of size 1. Our CAE had three, fully connected layers with 20442, 1000, and 50 parameters, resp. As noted below, we split the 60K networks into three subvectors to encode them with our autoencoder. The individual task-networks achieved accuracies ranging from 78\% to 84\%, and a mean accuracy of approximate 81\%.  Importantly, we encoded these larger networks using almost the same CAE architecture as the one used in the MNIST experiments. This was achieved by \textit{splitting} the 60K parameter vectors into three subvectors. As noted in Section~\ref{sec:methodology}, by splitting a larger input vector into smaller subvectors, we can encode networks of arbitrary sizes. As seen in Figure \ref{fig:combined_comparisons} (bottom), the accuracies of the reconstructed CIFAR networks also nearly matched the performances of their original counterparts, while also outperforming all other techniques.

\begin{figure}
    \centering
% 	\begin{center}	
		\includegraphics[width=0.9\linewidth]{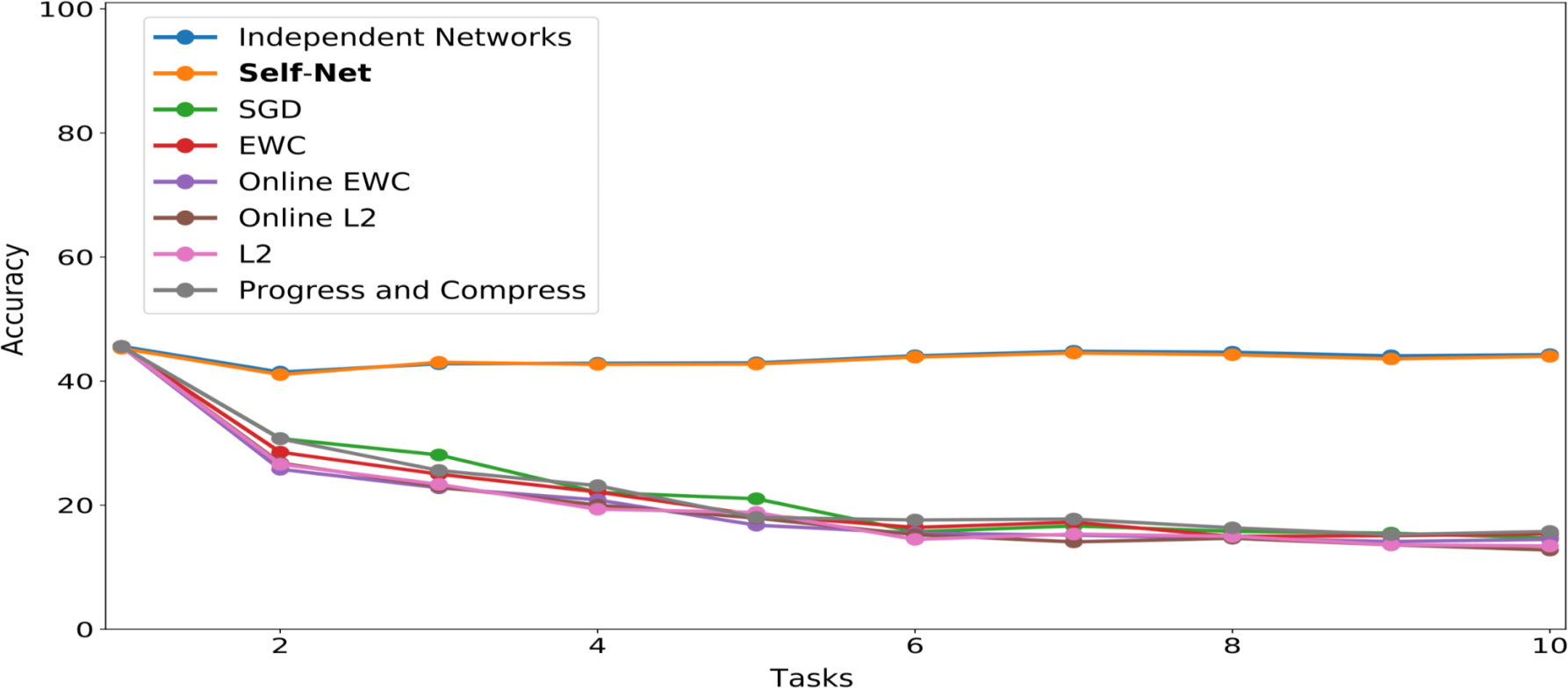}
% 	\end{center}
	\caption{CL performance comparisons with average test set accuracy on all observed tasks at each stage for CIFAR-100.}
	\label{fig:mean_accuracy_CIFAR100}
\end{figure}

% \subsection{Split CIFAR-100}
% \label{sec:cifar_100}

\textbf{Split CIFAR-100:} We applied the same learning approach for the CIFAR-100 dataset \cite{krizhevsky2009learning}. We split the dataset into 10 distinct batches comprised of 10 classes of images each. This resulted in 10 separate datasets, each designed for 10-class classification tasks. We used the same task-network architecture and Buffer size as in our CIFAR-10 experiments, modified slightly to accommodate a 10-class classification objective. The trained networks achieved accuracies ranging from 46\% to 59\%. We then encoded these networks using the same CAE architecture described in the previous experiments, again accounting for the input size discrepancy by splitting the task-networks into smaller subvectors.  As seen in Figure \ref{fig:mean_accuracy_CIFAR100}, our technique almost perfectly matched the performances achieved by independently trained networks.

\textbf{Incremental Atari:} To evaluate the CL performance of Self-Net in the challenging context of reinforcement learning, we used the code available at \cite{a3c_repo} to implement a modified Async Advantage Actor-Critic (A3C) framework, originally introduced in \cite{a3c_paper}, to attempt to learn successive Atari games while retaining good performance across all games. A3c simultaneously learns a policy and a value function for estimating expected future rewards. Specifically, the model we used was comprised of 4 convolutional layers (kernals of size 3x3, and strides of size 2x2), a GRU layer (800x256), and two ouput layers:  an Actor (256xNum\_Actions), and Critic (256x1), resulting in a complex model architecture and over 800K parameters.  Critically, this entire model can be flattened and encoded by the single AE in our Self-Net framework having three, fully connected layers with 76863, 2000, and 200 parameters, resp.  For these experiments we also used a Buffer of size 1.

\begin{figure}[t]
    \centering
	\includegraphics[width=0.9\linewidth]{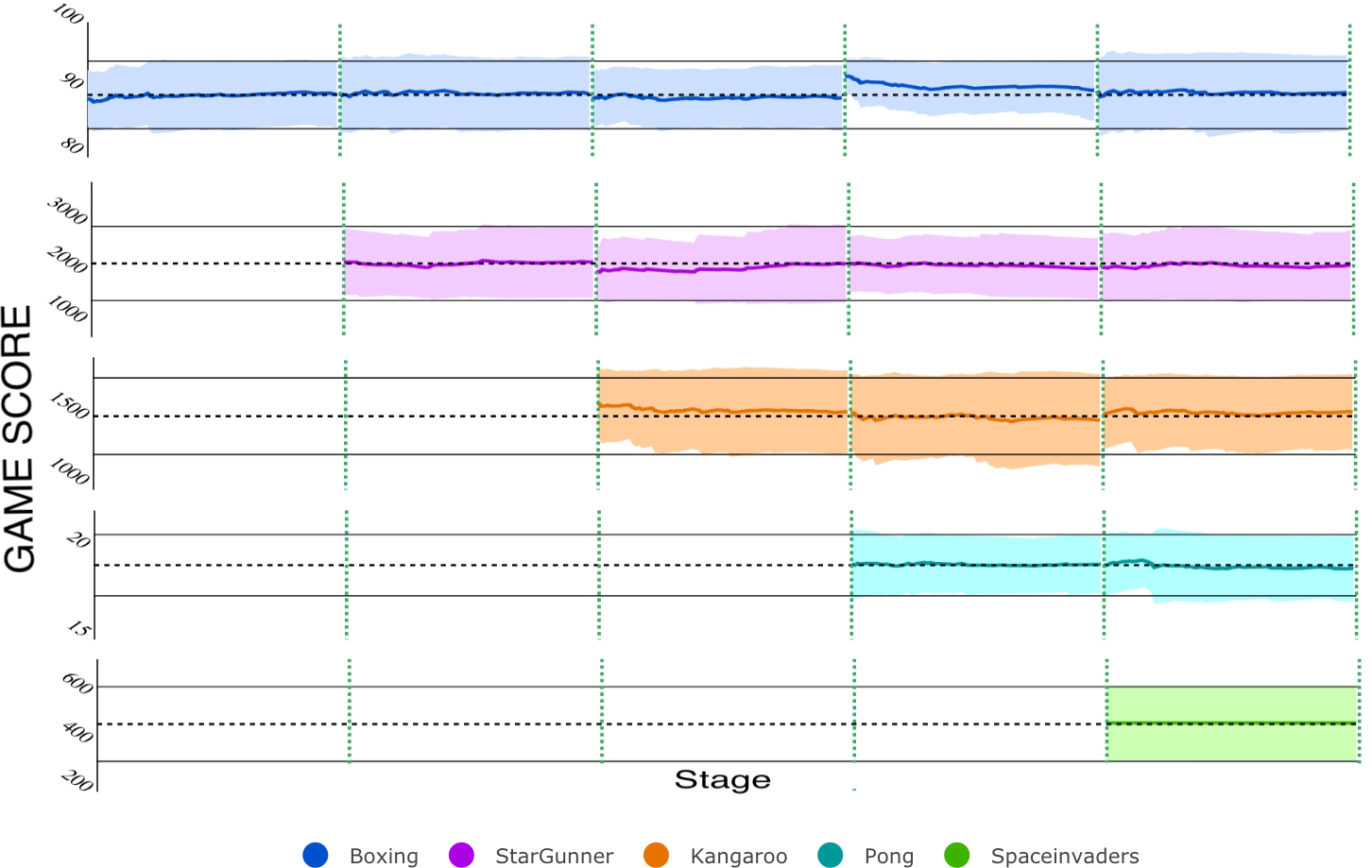}
	\caption{\textbf{CL on five Atari games with Self-Net:} To evaluate the reconstruction score at each stage, we ran the reconstructed networks for 80 full game episodes. The cumulative mean score is nearly identical to the original TN at each stage.}
	\label{fig:cl_atari}
\end{figure}

Similar to previous experiments, we trained our system on successive tasks, specifically the following Atari games:  Boxing, Star Gunner, Kangaroo, Pong, and Space Invaders. Figure~\ref{fig:cl_atari} shows the near-perfect retention of performance on each of the 5 games over the lifetime of the system. This was accomplished by training on each game only once, never revisiting the game for training purposes. The dashed, vertical lines demarcate the different stages of continual learning. That is, each stage indicates that a new  network was trained for a new game, over 40M frames.  Afterwards, the mean (dashed, horizontal black lines) and standard-deviation (solid, horizontal black lines) of the network's performance were computed by allowing it to play the game, unrestricted, for 80 episodes. After each stage, the performances of all reconstructed networks were examined by re-playing each game with the appropriate reconstructed network. As Figure~\ref{fig:cl_atari} shows, the cumulative means and SD's of the reconstructed networks closely mimic those achieved by their original counterparts.  

% During each stage, the AE was trained to integrate this new network while \textit{retaining} all previously-learned networks, as described in the preceding sections. 

% Significantly, even though the AE is retrained on \textit{recollections} of networks, it still retains superb performance, even in the challenging domain of Atari and, more broadly, reinforcment learning. To clarify, the network used to play the game of Boxing is recollected and reconstructed a total of 4 times during the course of these experiments. Similarly, the Star Gunner network was recollected 3 times; Kangaroo, 2 times; and Pong, 1 time.  That is, during the $5^{th}$, or rightmost stage in Figure \ref{fig:cl_atari}, the newly trained network for Space Invaders is in the process of being integrated into the latent space, but all 4 previously-learned networks are being retained using only their recollections, not the originally learned parameters.  Figure \ref{fig:cl_atari} shows how the AE retains a large amount of information while integrating a new network into its latent space.  The cosine-similarities of the previously-learned networks remain high, even while the AE learns to also reconstruct the new network.  

% \begin{figure}[H]
% \begin{center}

%   \includegraphics[width=0.99\linewidth]{atari_cosines_during_integration_5_skills.png}
% \end{center}
%   \caption{Integration of the 5th Atari network by the AE.  Note, the cosine-similarities of the 4 previously-learned networks remains high during retraining.}
% \label{fig:network_integration}
% \end{figure}

\subsection{Performance and storage scalability}
\label{sec:cl_large_tasks}
In CL, there is a trade-off between storage and performance. Using different networks for $k$ tasks yields optimal performance but uses $O(kn)$ space, while regularized methods such as Online EWC only require $O(n)$ space but suffer a steep drop in performance as the number of tasks grows. Our experiments on CL datasets show that our approach achieves much better performance retention than existing approaches by using slightly more space, $O(n + k\log{(n)})$. More precisely, we can quantify performance with respect to implicit storage compression. For example, by the tenth task, Online EWC \cite{HuszarE2496} has essentially performed 10x compression because it uses 1/10th of the overall storage required by ten different networks; however, its performance by this point is very poor. In contrast, our system achieves 10X compression when the size of the stored latent vectors grows to $10n$. In the following experiments, we verified that our method retains excellent performance even when reaching 10X compression, thus confirming that our AE is not simply memorizing previously learned weights.

\begin{figure}
    \centering
	\includegraphics[width=.78\linewidth]{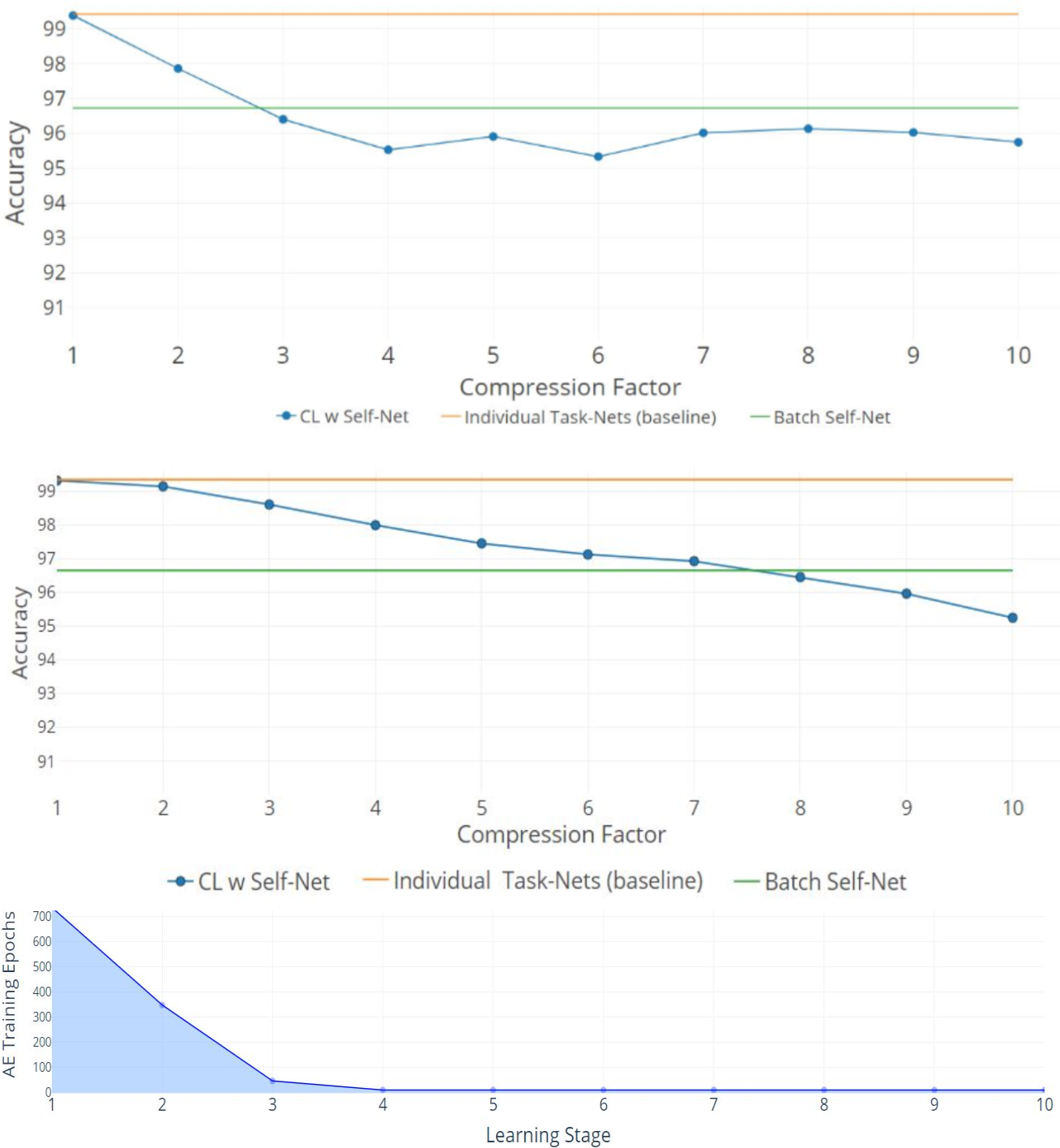}
	\caption{\textbf{10X Compression for Split-MNIST:} Orange lines denote the average accuracy achieved by individual networks, one per task.  Green lines denote the average accuracy when training the AE to encode all networks as a single batch.  Blue lines indicate the average accuracy obtained by Self-Net at each CL Stage. \textbf{Top:} 50 tasks with latent Vectors of size 5 and a Buffer of size 5. \textbf{Middle:} 100 tasks with latent vectors of size 10 and Buffer of size 10. The x-axis (top and middle) denotes the compression factor achieved at each learning stage. \textbf{Bottom:} the training epochs required by the 5-dimensional AE to incorporate new networks decreases rapidly over time.}
	\label{fig:selfNet_cl_combined}
\end{figure}

% across all previously encountered tasks

The top two plots of Fig.~\ref{fig:selfNet_cl_combined} show the mean performance for 50 and 100 Split-MNIST tasks, with latent vectors of size 5 and 10, resp. As before, the AE had 21432 input parameters. For comparison, we also plotted the original networks' performance and the performance of the reconstructions when the AE learns all the tasks in a single batch. The line with dots represents the CL system, where each dot indicates the point where the AE had to encode a new set of $m$ networks because the Buffer had filled up. For these experiments, we used a Buffer size $m$ of 5 and 10, resp.; these values were chosen so that each new batch of networks yielded an integer compression ratio, e.g., after encoding 15 networks with a latent vector of size 5, the Self-Net achieved 3X compression. Here, we fine-tuned all networks from the mean of the initial set of $m$ networks and penalized deviations from this source vector (using $\lambda = 0.001$), as described in Section~\ref{sec:methodology}. This regularization allowed the AE to incorporate subsequent networks with very little additional training, as seen in stages 4-10 (bottom image of Fig. \ref{fig:selfNet_cl_combined}).

% shows, once the AE has been trained on the first few batches of networks, during stages 1-3, the manifold is able to quickly adapt and incorporate subsequent networks with very little additional training, as seen in stages 4-10.  

For 10X compression, the Self-Net with a latent vector of size 5 retained $\sim$95.7\% average performance across 50 Split-MNIST tasks, while the Self-Net with 10-dimensional latent vectors retained $\sim$95.2\% across 100 tasks. This represents a relative change of only $\sim$3.3\% compared to the original performance of $\sim$99\%. In contrast, existing methods dropped to $\sim$50\% performance for 10X compression on this dataset (Fig.~\ref{fig:combined_comparisons}).

\subsection{Splitting networks and using multiple architectures}
\begin{figure}
    \centering
	\includegraphics[width=0.49\linewidth]{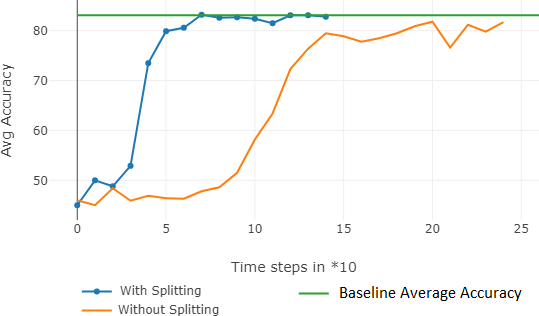}
	\includegraphics[width=0.49\linewidth]{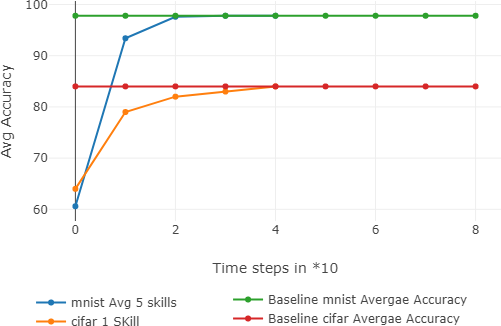}\\
	\caption{\textbf{Additional analyses:} \textbf{Left:} the AE training efficiency is improved when large networks are split into smaller subvectors. \textbf{Right:} a single AE can encode networks of different architectures and sizes.}
	\label{fig:split_multi_arch_combined}
\end{figure}

Splitting larger networks into smaller sub-vectors allows us to use a smaller AE. As an additional analysis, we verified that the smaller AE can be trained in substantially less time than a larger one. Figure~\ref{fig:split_multi_arch_combined} (left) shows the respective training rates of an AE with 20,000 input units (blue line)---trained to reconstruct 3 sub-vectors of length 20,000---compared to that of a larger one, with 61,000 input units (yellow line), trained on a 60K CIFAR-10 network. Clearly, using more inputs for a smaller AE enables us to more quickly encode larger networks. Finally, we also validated that the same AE can be used to encode trained networks of different sizes and architectures. Figure~\ref{fig:split_multi_arch_combined} (right) shows that the same AE can simultaneously reconstruct 5 MNIST networks and 1 CIFAR network so that all approach their original baseline accuracies.

\section{Conclusions and future work}
In this paper, we introduced a scalable approach for multi-context continual learning that decouples learning a set of parameters from storing them for future use. Our proposed framework makes use of state-of-the-art autoencoders to facilitate lifelong learning via continual self-modeling. Our empirical results confirm that our method can efficiently acquire and retain knowledge in continual fashion, even for very large numbers of tasks. In future work, we plan to improve the efficiency with which the autoencoder can continually model vast numbers of task networks. Furthermore, we will explore how to use the latent space to extrapolate to new tasks based on existing learned tasks with little or no training data. We also intend to compress the latent space even further (e.g., using only $\log{(k)}$ latent vectors for $k$ tasks). Promising approaches include clustering the latent vectors into sets of closely related tasks and using sparse latent representations. Finally, we will also investigate how to infer the current task automatically, without a task label.

\bibliographystyle{splncs04}
\bibliography{bibliography}
\end{document}